\documentclass[letterpaper,10pt,conference]{ieeeconf}

\IEEEoverridecommandlockouts

\usepackage{amssymb,amsmath}

\usepackage{algorithm}
\usepackage{algorithmic}
\usepackage{color}
\usepackage{xcolor}
\usepackage{wrapfig}

\makeatletter
\let\NAT@parse\undefined
\makeatother
\usepackage[sort&compress,numbers,sectionbib]{natbib}
\usepackage{booktabs}
\usepackage[small]{caption}
\usepackage{subcaption}
\usepackage{graphicx}
\usepackage{verbatimbox}
\usepackage{multirow}
\usepackage{balance}

\usepackage{tabularx}
\usepackage{xspace}
\usepackage{bm}
\usepackage{bbm} % bold math
\usepackage{pifont} % checkmark and X symbols
\usepackage{siunitx}
\newcommand{\figref}[1]{Fig.~\ref{#1}}
\newcommand{\tabref}[1]{Table~\ref{#1}}
\usepackage{hyperref}
\hypersetup{bookmarksopen,bookmarksnumbered,
pdfpagemode=UseOutlines,
colorlinks=true,
linkcolor=teal,
anchorcolor=teal,
citecolor=teal,
filecolor=teal,
menucolor=teal,
urlcolor=teal,
breaklinks=true
}
\newcommand{\xxnote}[3]{}
\ifx\hidenotes\undefined
  \usepackage{color}
  \renewcommand{\xxnote}[3]{\color{#2}{\textsf{#1: #3}}}
\fi
\usepackage[normalem]{ulem}
\useunder{\uline}{\ul}{}

\newcommand{\name}{{\textsc{Proact}}}

\newcommand{\tagline}{PROactive Assistance in Collaborative Transport}%{Coupled Object Motion Prediction and ConTrol}
\newcommand{\ee}{\mathrm{EE}}
\newcommand{\object}{\mathrm{obj}}
\newcommand{\nominal}{\mathrm{nom}}
\newcommand{\target}{\mathrm{target}}
\newcommand{\admittance}{\mathrm{adm}}

\newcommand{\workspace}{\mathcal{W}}
\newcommand{\obstacles}{\mathcal{O}}  % or \mathcal{R}?
\newcommand{\goal}{g}

\newcommand{\armdof}{N}

\newcommand{\pose}{x}
\newcommand{\twist}{\dot{\pose}}
\newcommand{\wrench}{w}
\newcommand{\action}{u}

\newcommand{\armPosition}{q}
\newcommand{\armVelocity}{\dot{\armPosition}}

\newcommand{\hist}{T}
\newcommand{\pred}{K}
\newcommand{\histRange}{{t-\hist+1 : t}}
\newcommand{\predRange}{{t+1 : t+\pred}}
\newcommand{\refPath}{\mathcal{P}^{xy}}

\title{
Learning from Humans for Proactive Assistance\\ in Human-Robot Collaborative Transport %Physical Human-Robot Collaboration
}

\author{
  Elvin Yang and Christoforos Mavrogiannis\thanks{The authors are with the Department of Robotics, University of Michigan, Ann Arbor, United States. Email: $\{$eyy, cmavro$\}$@umich.edu}
}

\begin{document}
\maketitle

%===============================================================================

\begin{abstract}

% "these objectives" - what objectives?
%
% In the related work, we argue two classes of prior work:
% 1. papers to model/predict future human/team behavior
% 2. papers to achieve whole body control, compliant control, often both.
%
% Why are each of these two classes necessary? What downstream behavior do they enable?
%
% - reduction of human effort
%     - proactive behavior to achieve task completion
%     - not "getting in the way"
% - coordination with the human ; harmony ; congruous ; concordant ; removing discordance
%     - avoid overriding, overpowering/pushing the human
%     - does this also count as reducing the human's effort?

% Prior work tends to isolate these objectives;
%
% - to enable more efficient physical relocation and reduce human effort, works propose models of collaborative human behavior, but tend to evaluate on platforms without manipulability (including limiting the manipulability by locking the arm) and at low speeds.
% - to maintain smooth physical interaction, works develop compliance mechanisms, but evaluate them in simple settings, e.g., straight lines, environments without obstacles, freely moving around without a clear navigational objective.

% =============

We focus on human-robot collaborative transport, a challenging task of broad relevance spanning logistics, manufacturing, and the home, in which a user and a robot work together to relocate a large or heavy object.
% This needs to stay consistent with metrics to motivate what we actually did.
To act as an effective partner, the robot should reduce the user's effort by contributing to efficient relocation of the object while remaining physically responsive to them.
% without imposing unilateral motion/decisions/control over the object.
% avoid forceful overriding, overpowering, dragging, or pushing the human.
% maintaining harmony/agreement/concordance/conccurrence 
%
% This is now more of a logical argument than an empirical one.
% Prior work tends to isolate these objectives, resulting in robots that either override user input or require the user to continuously direct the robot.
Prior work often addresses these capabilities separately, producing robots that may move the object efficiently but resist user input, or accommodate the user but depend on continuous guidance.
% Prior work tends to study these objectives separately, either simplifying robot execution for task-level modeling or simplifying the task setting for compliant control.
% \cmnote{we need to make this point more clear, explain it with simpler words}
% \elvin{changed in intro, double check to match} Our key insight is that integrating predictions of human collaborative behavior with compliant robot control can enable the robot to proactively assist with transporting the object while remaining responsive to physical input from the user.
Our key insight is that obstacle-constrained collaborative transport requires integrating predictions of human collaborative behavior with compliant robot control.
To this end, we introduce \name{}, a framework for human-robot collaborative transport that incorporates anticipation into compliant whole-body control through a learned model of human collaborative behavior.
Trained on a large-scale, real-world dataset of dyadic human transport demonstrations, our transformer architecture distills collaborative behavior into predictions of future object motion.
Across 108 real-world trials with a 9-DoF mobile manipulator, \name{} reduces mean interaction work by 59.2\% and 20.4\%, and mean completion time by 12.9\% and 6.9\%, relative to compliance-only and MPC baselines, respectively. Footage from our experiments can be found at~\url{https://youtu.be/qAGvQfVPjbk}.

% improves transport efficiency (12.9\%, 6.9\%) while simultaneously reducing physical interaction effort (59.8\%, 21.4\%) relative to both a compliance-only controller and a conventional MPC baseline, respectively.
% reduces interaction forces, torques, and work relative to both a compliance-only controller and a conventional MPC baseline.
\end{abstract}

% \begin{IEEEkeywords}
% Human-Robot Interaction, Imitation Learning, Mobile Manipulation
% \end{IEEEkeywords}

\section{Introduction}

% increased interest in human-robot collaboration, particularly to develop robots capable of seamlessly assisting people with physically demanding tasks.
% human-robot interaction has received increasing attention for enabling robots to assist people in everyday life.
% As robots become more widely accessible, there is increasing interest in deploying them in factories, logistics, and home environments.
% In these settings, many useful forms of assistance require robots to physically collaborate with humans rather than merely operating nearby or in isolation.
% Despite advances in robot hardware and algorithmic modeling, elevating robots to function as effective partners in physically demanding tasks remains a long-standing goal.
% \cmnote{these 3 first sentences are quite generic and dont convey much, i would remove}

% \cmnote{you might want to add 1 sentence to introduce the relevance of physical HRI as a critical skill that may enable robots to serve as valuable partners across important domains, then in 2nd sentence you say that collab transport is one such task}

% Physical human-robot interaction is a central skill for robots to function as useful task partners as they become more widely deployed in factories, logistics, and home environments.
% As robots become more widely deployed in factory floors, warehouses, and homes, physical human-robot interaction is central to their ability to function as useful partners.

Collaborative transport, in which two agents jointly maneuver an object through an environment, has emerged as a representative setting for developing and evaluating physical human-robot interaction~\citep{kosuge2000mobile,stuckler2011following,noohi2016model}.
This task is challenging because an effective robot partner should aim to reduce the user's physical effort~\citep{lima2023assistive,shao2024constraint,du2025learning,solak2025context} not only by supporting the object's weight, but also by contributing to functional aspects of the task, such as goal progress, obstacle avoidance, and object stability.
At the same time, the robot should account for the user's intent and respond to physical input to avoid unilaterally dictating the object's motion.
% These challenges have led to two recurring lines of work: modeling human-intended goals, trajectories, and collaborative strategies, and executing motion through compliant and whole-body control.
These considerations have led to two main lines of work.
One addresses immediate interaction dynamics through compliant and whole-body control, enabling the robot to react to physical input from the user while coordinating many degrees of freedom~\citep{agravante2019human,du2025learning}.
The other focuses on human modeling, in the form of intended goals, trajectories, and task strategies, providing the robot with context beyond immediate physical cues~\citep{rysbek2024proactive,ng2023takes,yang2025implicit}.
% allowing the robot to anticipate how the task may unfold
% Directly executing commands without compliance
% However, each line of work often abstracts the other side of the problem: control approaches may simplify the task setting by assuming obstacle-free environments, straight-line motion, or no explicit target destination, whereas human-modeling approaches are often demonstrated on robot systems with limited speed and manipulability.
However, these aspects are often studied in isolation, limiting the robot's effectiveness as a collaborative partner: purely compliant systems leave the user fully responsible for task progression, while human models are often deployed without preserving the user's influence over the carried object.
As a consequence, prior evaluations frequently restrict the demonstrated scope to obstacle-free environments or straight-line motion, or use systems that substantially constrain interaction speed or object motion.

% \cmnote{could you give more intuition about why these methods make these assumptions? what is their fundamental limitation that necessitates these assumptions?} \elvin{It may be a wording issue on my part, but the scope of the paper would be defined in a simplified setting, e.g., an admittance or whole-body controller evaluated in back-and-forth motion in a wide open space, with no explicit representation or reasoning of obstacles or a goal location.}, 
% often simplify robot execution by using low speeds or limiting object manipulability \cmnote{again, we need to be more clear about why each class of methods makes these assumptions}.
% These limit the practical usability of the system in real collaboration.

% \cmnote{i think one important contribution of this work is the richer modeling of human intent over a multistep horizon, rather than adopting a purely reactive response to the current input of a 1-step horizon. this is currently not highlighted}

% \cmnote{
% i think we should be clear that we learn a model of collaborative human behavior; the learned object dynamics reflect how the object moves in response to dyadic human collab transport; by modeling this dynamics, we can make the robot capable of anticipation
% implicit assumption we make that we should note somewhere: the human partner works with the robot similar to how they would work with a fellow human
% }

\begin{figure}[t]
    \centering
    \includegraphics[width=\linewidth]{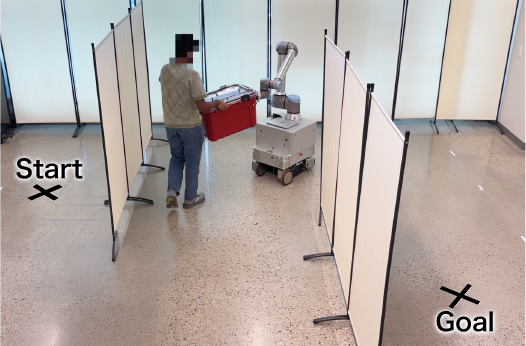}
    \caption{
        Real-world collaborative transport with a user and a 9-DoF mobile manipulator navigating an obstacle-constrained workspace. The robot runs \name{}, combining a learned model of human collaborative behavior with compliant whole-body control to provide proactive assistance during physical interaction.
    }
    \label{fig:figure1}
\end{figure}

Our key insight is that effective collaborative transport through obstacle-constrained environments requires a combination of long-horizon predictions of collaborative human behavior and compliance.
To this end, we introduce \textit{\tagline} (\name).
The core of \name{} is a model of collaborative human behavior that anticipates how the collaborative interaction may unfold given a history of past interaction.
We learn this model from a large-scale dataset of dyadic human collaborative transport~\citep{freeman2024classification}, using object motion as a compact representation of collaborative interaction.
We integrate these predictions with a compliant whole-body controller, which we deploy on a 9-DoF mobile manipulator. This integration enables the robot to interact proactively while adapting to physical user input, resulting in more efficient human-robot collaboration (HRC) compared to baselines. While our framework was demonstrated on collaborative transport, its implications and relevance transfer to the broader space of physical HRC, including collaborative lifting, placement, and assembly.

In summary, our contributions include:
\begin{itemize}
    \item A transformer-based model for human collaborative behavior prediction, learned from real-world demonstrations of human dyadic transport.
    \item A closed-loop system integrating collaborative behavior prediction with compliant whole-body control for mobile manipulation on high-DoF platforms.
    % Validation of prediction accuracy 
    \item Model validation on an independent test set and real-world system integration and evaluation using a 9-DoF mobile manipulator in an experimental setting of practical relevance.
    % collaborative performance across 108 real-world human-robot trials in an obstacle-constrained workspace.
    % \item Extensive demonstrations of the integrated system on a 9-DoF mobile manipulator in an experimental setting of practical relevance. 
\end{itemize}

\begin{figure*}[t]
    \centering
    \includegraphics[width=\linewidth]{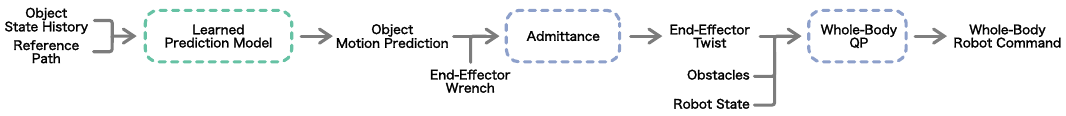}
    \caption{
        Architecture of \name{}. A learned prediction model maps object state history and a reference path provided by a global planner to future object motion, which is adapted by an admittance layer and tracked by a whole-body quadratic program (QP).
    }
    \label{fig:architecture}
\end{figure*}

\section{Related Work}

\textbf{Physical HRC}. Physical HRC~\citep{selvaggio2021survey} has received substantial attention in recent years. Much of the prior work emphasizes comanipulation between humans and robots in static environments where the manipulator base is fixed~\citep{zheng2023safe,gienger2018human,shao2024constraint,peternel2017human}. When base mobility is involved, it tends to involve short-horizon or point-to-point motion~\citep{ranatunga2015intent,takubo2002human,shao2024constraint}, straight-line segments~\citep{lima2023assistive,sirintuna2022human,bussy2012human}, isolated planar translation and rotation~\citep{mielke2024human}, or manipulation without a clearly defined goal~\citep{du2025learning} in obstacle-free environments. Typical architectures involve whole-body control via quadratic programs~\citep{sirintuna2022human,agravante2019human,kumbhar2025mpcqp} or reinforcement learning~\citep{cao2026paint,du2025learning}. The robot often assumes a rigid follower role, directly responding to human intent via compliant control~\citep{lima2023assistive,solanes2018human,takubo2002human,stuckler2011following,sirintuna2022human,mielke2024human,kosuge2000mobile}. Online adaptive admittance and impedance methods extend compliance by updating controller parameters during interaction based on human motion, EMG signals, or force-based intention estimates~\citep{shao2024constraint,peternel2017human,ranatunga2015intent}. However, these methods still require continued human effort to initiate and guide the robot's motion and often assume instrumented objects~\citep{mielke2024human,bussy2012proactive,bussy2012human,gienger2018human}, which result in high design complexity and limit generalization. Our approach builds on quadratic programming for whole-body robot control and admittance-based compliance but, in contrast to prior work which relegates the robot to a reactive follower, we incorporate object-level collaborative dynamics to enable proactive robot behavior generation. We move beyond prior work in terms of experimental settings, demonstrating human-robot collaborative transport on a 9-DoF mobile manipulator at human-like speeds through an obstacle-occupied workspace.

\textbf{Human models for physical HRC}. Human modeling in physical HRC~\citep{hoffman2024inferring} is often cast as prediction of sequences of low-level signals, such as forces~\citep{ng2023diffusion,dominguez2023improving,dominguez2024exploring,dominguez2024force} and velocities~\citep{dominguez2024exploring,dominguez2024force,mielke2024human,townsend2017estimating}, or as inferring higher-level objectives, such as the intended goal~\citep{rysbek2024proactive,rysbek2023recognizing,shao2024constraint}, future object trajectories~\citep{ng2023takes}, or workspace traversal strategies~\citep{yang2025implicit}. Human and robot behavior in physically collaborative tasks is coupled, motivating recent work in collaborative behavior modeling~\citep{ng2023takes,mielke2024human}. Part of the challenge in developing rich models of collaborative human behavior lies in the lack of appropriate datasets~\citep{youssef2017uehri,yasar2024posetron,celiktutan2019mhhri,kratzer2020mogaze,newman2022harmonic,tian2023crafting}: some are missing important modalities like 3D pose and environment representations~\citep{freeman2024classification}, whereas others lack physical interactions between agents~\citep{tian2023crafting,yasar2024posetron}. Here, we leverage the dataset of~\citet{freeman2024classification}, which captures diverse 6-DoF object motions from transport demonstrations of human dyads, to train a state-of-the-art transformer architecture modeling human-robot collaborative dynamics. Similar to the work of~\citet{ng2023takes} and~\citet{mielke2024human}, we cast the problem of collaborative behavior modeling as estimation of the carried object's future motion. Our work differs in that it incorporates high-fidelity predictions over longer horizons, by building upon a state-of-the-art transformer architecture. Additionally, the predictions themselves serve as nominal commands for compliant whole-body control on a 9-DoF mobile manipulator, enabling proactive adaptation to expected collaborative human behavior.

\section{Problem Statement}

% Formalize the problem we are trying to solve in a way that is decoupled from the solution, using only absolutely necessary notation. Conclude with a sentence stating what exactly we are trying to accomplish.

% while jointly maintaining control of the object at all times, avoiding drops and slips
We consider a user and a robot collaborating to transport a rigid object in a workspace $\workspace \subseteq SE(3)$ containing obstacle-occupied regions $\obstacles \subset \workspace$.
% \mathfrak{se}(3)
We represent the object state as $(\pose^\object, \twist^\object) \in \workspace \times \mathbb{R}^6$, where $\pose^\object$ is its pose and $\twist^\object$ is its twist.
The user and robot maintain fixed grasps on opposite sides of the object throughout the task.
The robot is a mobile manipulator with a holonomic base and an $\armdof$-DoF arm.
% the robot executes a control action $\action \in \actionSpace$ at every time step.
% move -> transport, carry?
% semantically, "carrying" intentionally excludes sliding/rolling the object on the floor as a mode of transport, which would result in friction effects
The objective of the team is to transport the object from an initial state $(\pose^\object_0, \mathbf{0})$ to a goal state $(\pose^\object_\goal, \mathbf{0})$ while avoiding collisions with $\obstacles$.
Our goal is to enable the robot to efficiently transport the object with a user while avoiding collisions with the static obstacles and the user, maintaining smooth object motion, and requiring low physical interaction effort from the user.
We assume the robot has access to an end-effector force-torque sensor, object state history, the workspace layout, and a global path planner.

\section{\name{}: A Framework for Proactive Assistance in Collaborative Transport}

% hierarchical?
We propose a framework that combines human collaborative behavior modeling in the form of object motion prediction with compliant whole-body robot control for physical human-robot collaborative transport.
The model, learned from demonstrations of dyadic human collaborative transport, predicts a future horizon of object twists describing how a collaborative team would realize a route, conditioned on recent object state history.
At each model update during deployment, the first twist in the predicted horizon serves as a nominal object motion target for the whole-body controller.
An overview of our framework is shown in~\figref{fig:architecture}.

% apply the model in a receding-horizon manner and use the first predicted twist as a nominal object motion target for the whole-body controller.

\subsection{Learning a Model of Object Twist Prediction}

Our goal is to learn how a team is likely to move the object next based on how the team collaborated in the near past and the route the team is following.
% Intuitively, our goal is to learn a model that predicts how an object is likely to move next, given observations of how it moved in the near past as a result of dyadic human collaborative behavior.
We formulate this problem as object twist prediction over a finite horizon:
\begin{equation}
    \hat{\twist}^{\object}_{\predRange}
    = F \left(
        \pose^\object_{\histRange},
        \twist^\object_{\histRange},
        \refPath_t
    \right)\mbox{,}
\end{equation}
where at time $t$, $\hat{\twist}^{\object}_{\predRange}$ are predicted object twists over the next $\pred$ time steps, $\pose^\object_{\histRange}$, $\twist^\object_{\histRange}$ are $\hist$ steps of object state history, and $\refPath_t$ is a local reference path for the object.
% where at time $t$, the model $F$ predicts object twists over the next $\pred$ time steps $(\hat{\twist}^{\object}_{\predRange})$ from $\hist$ steps of object state history $(\pose^\object_{\histRange}$, $\twist^\object_{\histRange})$ and a local reference path $(\refPath_t)$.
The reference path is a geometric route with no timing, velocity, or orientation information, and may, for example, be supplied by a global planner.
The model $F$ predicts a 6-DoF twist sequence describing how a collaborative team would move the object along the route.
% The model $F$ predicts how a collaborative team would manipulate the object to traverse it as a time-varying 6-DoF twist sequence.

% The high-level idea from the "path" input to to the model is that there is no guarantee that the human-robot team will execute it precisely. Rather, it serves as a "high level" signal to the robot about how the object may be manipulated. In our case, the path is quite low dimensional since it has no vertical or angular information.

% \begin{figure}[t]
%     \centering
%     \includegraphics[width=\linewidth]{figures/data-global-path.png}
%     \caption{Object motion capture traces (blue) from 522 trials and average global path (black). \cmnote{is this from your experiments? from freeman? be clearer. also make sure any figure fonts are legible}}
%     \label{fig:data-global-path}
% \end{figure}

\textbf{Model Architecture}.
We instantiate the model as a transformer encoder~\citep{vaswani2017attention}.
Separate linear projections map each 13-dimensional object state in the history and each 2D point in $\refPath_t$ to 256-dimensional token embeddings.
We append $\pred$ learned query tokens, one for each future prediction step, and add separate learned positional embeddings to the history, path, and query tokens.
The resulting sequence is processed by four transformer encoder layers with eight self-attention heads and 512-dimensional feedforward layers.
A linear head maps the encoded query tokens to produce the 6-DoF object motion prediction $\hat{\twist}^{\object}_{\predRange}$.

\textbf{Training}.
We use 6-DoF object trajectory data from the collaborative transport dataset of~\citet{freeman2024classification}.
The dataset contains human-human transport trials from 31 participant pairs, spanning multiple obstacle and behavioral instruction conditions.
In each trial, participants transport a \qty{27}{\kilogram} stretcher-like object with dimensions \qtyproduct{1.220 x 0.602 x 0.167}{\meter} while holding opposite ends. We downsample trajectories from \qty{200}{\hertz} to \qty{20}{\hertz} to reduce noise when computing object twists and to set the model update rate for deployment.
We compute object twists from consecutive poses using the $SE(3)$ logarithm map and express them in the object body frame.
We then segment each trajectory into windows containing $\hist$ history steps and $\pred$ future steps. Because the dataset records demonstrations along prescribed routes but does not provide obstacle geometry, we construct a geometric reference path, represented as 2D points spaced \qty{0.1}{\meter} apart, from an average of the demonstrated object paths.
For each trajectory window, $\refPath_t$ is a local segment of the reference path near the object's current position. We express both the object pose history and $\refPath_t$ in the current object frame. We split the dataset into training and validation sets by participant pair in a 75/25 ratio.
We train the model for 50 epochs using Smooth L1 (Huber) loss, AdamW, and an initial learning rate of $10^{-3}$. We retain the checkpoint with the lowest validation loss.

\subsection{Low-Level Robot Control}

Because we assume that the robot maintains a fixed grasp on the object, we transform the first predicted object twist $\hat{\twist}^{\object}_{t+1}$ to its end-effector $\twist^{\ee,\nominal}_{t+1}$.
Low-level execution then proceeds in two stages: an admittance layer modifies the nominal end-effector twist, and a whole-body QP controller maps the resulting twist to arm and base velocities.

\textbf{Admittance Layer}. To support compliant physical interaction, the admittance layer adjusts the nominal end-effector twist based on wrench measurements at the robot end effector.
The admittance layer maintains a virtual velocity state $v^{\admittance}_t \in \mathbb{R}^6$ in the end-effector frame:
\begin{align}
    v^{\admittance}_0 &= \mathbf{0}, \\
    v^{\admittance}_{t+1}
    &=
    M^{-1}(\tilde{\wrench}^{\ee}_t - D v^{\admittance}_t)\Delta t
    + v^{\admittance}_t, \label{eq:admittance-update}
\end{align}
where $M,D\in\mathbb{R}^{6\times6}$ are diagonal virtual mass and damping matrices, $\tilde{\wrench}^{\ee}_t \in \mathbb{R}^6$ is an end-effector wrench measurement after a low-pass filter and deadband are applied to attenuate high-frequency noise and sensor bias, and $\Delta t$ is the low-level control timestep.
We add the admittance velocity to the nominal end-effector twist to produce a target end-effector twist for the whole-body controller:
\begin{equation}
    \twist^{\ee,\target}_{t+1}
    =
    \twist^{\ee,\nominal}_{t+1} + v^{\admittance}_{t+1}.
\end{equation}
% We omit a stiffness term in the admittance layer so it does not impose virtual-spring position correction.
In the absence of external wrench input, the admittance velocity decays toward zero.

\textbf{Whole-body Robot Control}. The whole-body controller tracks the target end-effector twist $\twist^{\ee,\target}_{t+1}$ by optimizing a robot control action $\action_t$ comprising arm joint velocities $\armVelocity_t \in \mathbb{R}^\armdof$ and base velocity
$\begin{bmatrix}
    v_{x,t} & v_{y,t} & \omega_{z,t}
\end{bmatrix}$.
We formulate the controller as the following quadratic program:
\begin{align}
    \underset{\action_t,\varepsilon,\alpha}{\min}\ %
        & \Vert W_u \action_t \Vert_2^2 \label{eq:qp-cmd-mag} \\
        & + \Vert W_s (\action_t - \action_{t-1}) \Vert_2^2 \label{eq:qp-cmd-smooth} \\
        & + \Vert W_p (\armVelocity_t - \armVelocity^{\mathrm{posture}}_t) \Vert_2^2 \label{eq:qp-cmd-posture} \\
        & + \Vert W_\varepsilon \varepsilon \Vert_2^2 \\
        & - w_\alpha \alpha \\
    \text{s.t.}\quad
        & J_t \action_t - \alpha \twist^{\ee,\target}_{t+1} - \varepsilon = 0 \label{eq:qp-twist_track}\\
        & \action_{\min} \le \action_t \le \action_{\max}\label{eq:qp-ulimits} \\
        & \lVert [v_{x,t}, v_{y,t}]^\top \rVert_2 \le v_{\max} \label{eq:qp-base_speed} \\
        & |\varepsilon_i| \le \varepsilon_{\max,i} \\
        & \alpha \in [0,1] \\
        & n_i \cdot \begin{bmatrix}
            v_{x,t} \\
            v_{y,t}
          \end{bmatrix}
          \ge -\kappa h_i \label{eq:qp-obstacles}
        % & A_{\obstacles}(q_t)\action_t \ge b_{\obstacles}(q_t), \\
\end{align}
% where $\Vert W_u \action_t \Vert_2^2$ penalizes large robot commands and discourages unnecessary arm or base motion;
% $\Vert W_s(\action_t-\action_{t-1})\Vert_2^2$ penalizes command changes between control cycles; 
% $\Vert W_p(\dot{q}_{\mathrm{arm},t}-u^{\mathrm{posture}}_t)\Vert_2^2$ biases the redundant arm toward a nominal configuration, where $u^{\mathrm{posture}}_t$ is a clipped proportional velocity command computed from the arm joint-position error;
% $\varepsilon \in \mathbb{R}^6$ allows bounded deviation from the target twist to reduce infeasibility;
% and the scalar $\alpha \in [0,1]$ scales the target twist when the full command cannot be achieved, preserving the direction of the target twist rather than clipping individual twist components.
where \eqref{eq:qp-cmd-mag}, \eqref{eq:qp-cmd-smooth}, and \eqref{eq:qp-cmd-posture} regularize command magnitude, command smoothing, and redundant arm posture, respectively.
$\armVelocity^{\mathrm{posture}}_t$ is a bounded joint-velocity command that drives the arm toward a nominal posture.
The slack variable $\varepsilon \in \mathbb{R}^6$ allows bounded deviation from the target twist to reduce solver infeasibility.
The scalar $\alpha \in [0,1]$ uniformly scales the target twist, preserving the direction of the motion when needed to stay within kinematic limits.
Eq.~\eqref{eq:qp-twist_track} enforces twist tracking through the whole-body Jacobian $J_t$, evaluated at the current robot configuration, up to the slack $\varepsilon$ and scale factor $\alpha$.
Eq.~\eqref{eq:qp-ulimits} enforces limits on individual arm and base velocity components, and eq.~\eqref{eq:qp-base_speed} enforces a planar base speed limit via polygonal approximation.
We represent workspace boundaries and obstacles as collections of line segments.
Eq.~\eqref{eq:qp-obstacles} adds obstacle avoidance through linear constraints on the planar base velocity.
For each nearby line segment $i$, $n_i$ is the unit vector from the closest point on the segment to the base, $h_i$ is the clearance from the base to the segment, and $\kappa>0$ controls the maximum allowed speed toward the segment as a function of clearance.

\section{Evaluation}\label{sec:experiments}

\begin{figure}
    \centering
    \includegraphics[width=\columnwidth]{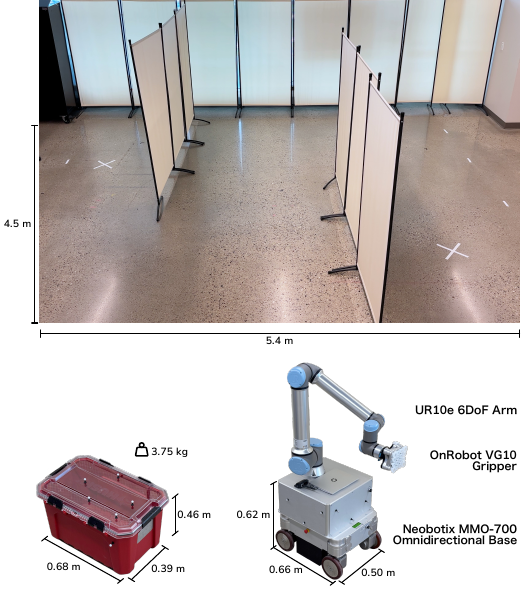}
    \caption{Evaluation setting with an obstacle-constrained workspace, object, and mobile manipulator. X-shaped floor markings indicate start and goal locations.}
    \label{fig:setting}
\end{figure}

% model-level, algorithm-level, prediction-level
We conduct two complementary evaluations of our method: a component-level evaluation of model prediction performance and a pilot study assessing the full system via real-world human-robot collaborative transport trials.
Motivated by practical transport challenges such as maneuvering around corners and through constrained spaces, we design a new \qtyproduct{4.5 x 5.4}{\meter} workspace that requires successive turns between start and goal locations (\figref{fig:setting}).
% Both evaluations take place in a new \qtyproduct{4.5 x 5.4}{\meter} workspace whose obstacle arrangement differs substantially from that of the training environment.
% Motivated by practical transport challenges such as maneuvering around corners and through constrained spaces, the workspace layout requires successive turns between the marked start and goal poses (\figref{fig:setting}).
An OptiTrack motion capture system provides robot and object poses within the workspace at \qty{120}{\hertz}.
% Readily available (in North America as of May 2026): https://www.homedepot.com/p/311485319
% The box itself is measured as 3.47 kg. With the steel plate for better grip surface, the box is measured as 3.75 kg.
% These dimensions are taken from the product label. Noteworthy that the dimensions are of the rectangular bounding box. If you multiply the dimensions you get 122 L, but the inner volume of the box is 75.7 L (20 US Gal). The box is also a slight frustum.
For the transported object, we use a commercially available heavy-duty storage container with dimensions \qtyproduct{0.678 x 0.388 x 0.464}{\meter} and a mass of \qty{3.75}{\kilogram} (\figref{fig:setting}). Footage from our experiments can be found at~\url{https://youtu.be/qAGvQfVPjbk}.

\subsection{Model Prediction Performance}

\begin{figure}[t]
    \centering
    \includegraphics[width=\columnwidth]{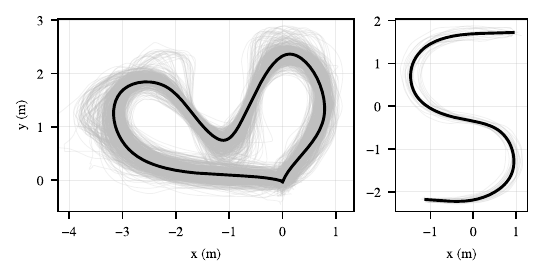}
    \caption{We train our transformer-based object motion prediction model on a set of object trajectories drawn from the dataset of~\citet{freeman2024classification} (left), and test model performance on a substantially different set collected in our lab (right). Gray curves represent object paths from individual trials, whereas black curves indicate averaged paths.}
    \label{fig:reference-paths}
\end{figure}

% TODO(elvin): We can bump this to 3-4 pairs and 30-40 trials to be able to draw more generalizable conclusions from the test evaluation.
We collect an independent test dataset containing 22 human-human transport trajectories in the evaluation environment with two pairs of members from our research group.
In each trial, motion capture records the object pose while users grasp it from opposite ends and carry it between start and goal poses marked on the floor.
% \cmnote{what kind of motion/trajectories did you collect? maybe an overlay of the object traj data is good (perhaps contrasting against a similar figure from freeman traj data could be informative?)}
We preprocess the test dataset using the same procedure as the training data.
\figref{fig:reference-paths} compares trajectories from the training and test datasets, illustrating the different workspace layouts.
We evaluate multiple horizon lengths $\pred$ to characterize the tradeoff between longer-horizon anticipation and prediction accuracy.
For each value of $\pred$, we train a separate model and tune it on the validation set.

\begin{figure*}[t]
    \centering
    \begin{subfigure}{0.49\textwidth}
        \centering
        \includegraphics[width=0.95\linewidth]{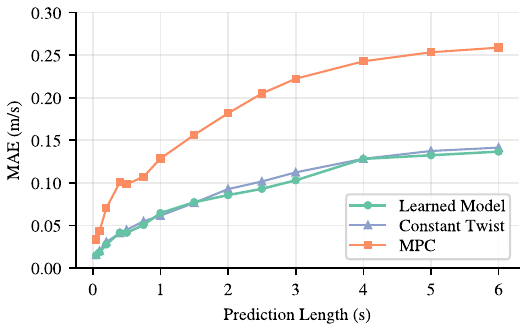}
        \caption{Translational twist components.}
        \label{fig:model-perf-test-linear}
    \end{subfigure}
    \hfill
    \begin{subfigure}{0.49\textwidth}
        \centering
        \includegraphics[width=0.95\linewidth]{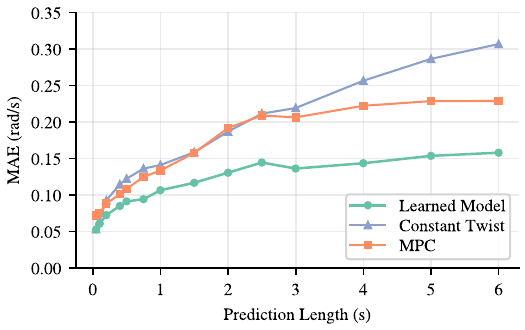}
        \caption{Rotational twist components.}
        \label{fig:model-perf-test-angular}
    \end{subfigure}

    \caption{Test-set prediction error for translational (\subref{fig:model-perf-test-linear}) and rotational (\subref{fig:model-perf-test-angular}) twist components across a range of prediction lengths.}
    \label{fig:model-perf-test}
\end{figure*}

\textbf{Baselines}.
We evaluate the learned model on the test dataset against future object twist sequences produced by the following algorithms:
\begin{itemize}
    \item \emph{Constant Twist}: propagates the most recent object twist, i.e., $\hat{\twist}^{\object}_{\predRange} = \twist^{\object}_{t}$, generalizing the constant-velocity baseline that tends to deliver practical performance in short-term motion prediction~\citep{scholler2020constant,salzmann2020trajectron++}.
    % \item \textbf{Constant Acceleration}: extrapolates from the most recent twist using the average acceleration over the history window. \elvin{Removed: this was a "try it because it might be interesting prediction baseline", but it does not perform well. It would only make sense for very small K and doesn't seem to be used by prior work.}
    \item \emph{Model Predictive Control (MPC)}: optimizes object twists over a receding horizon using a cost combining path tracking, smoothness, and goal progress, representing conventional goal-directed motion planning.
    % but does not model dyadic collaborative behavior.
\end{itemize}

% \textbf{Metrics}.
% We report mean absolute error (MAE) over the prediction horizon.
% and MAE for the first predicted twist.
% The first-step MAE is relevant for deployment because the controller uses the first prediction in a receding-horizon manner before replanning at the next control cycle.

\begin{figure*}[t]
    \centering
    \begin{subfigure}{0.85\textwidth}
        \centering
        \includegraphics[
            trim={0cm 0.3cm 0cm 0.2cm},
            clip=true,
            width=\linewidth
        ]{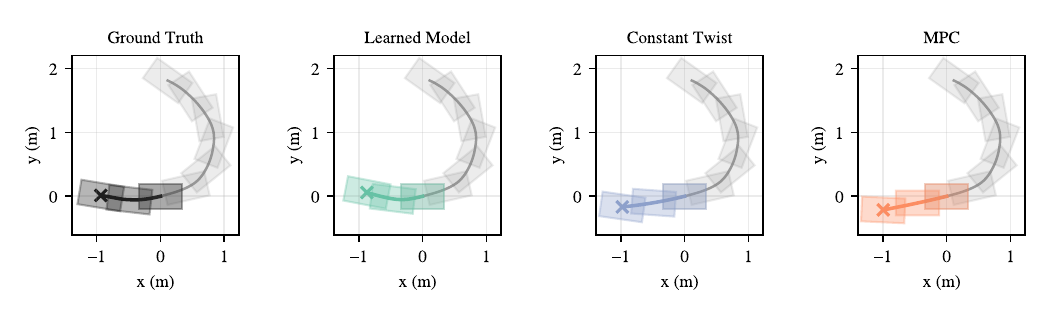}
        \caption{Motion rollouts produced by each algorithm.}
        \label{fig:model-rollouts}
    \end{subfigure}

    \begin{subfigure}{0.85\textwidth}
        \centering
        \includegraphics[
            trim={0cm 0.27cm 0cm 0.2cm},
            clip=true,
            width=\linewidth
        ]{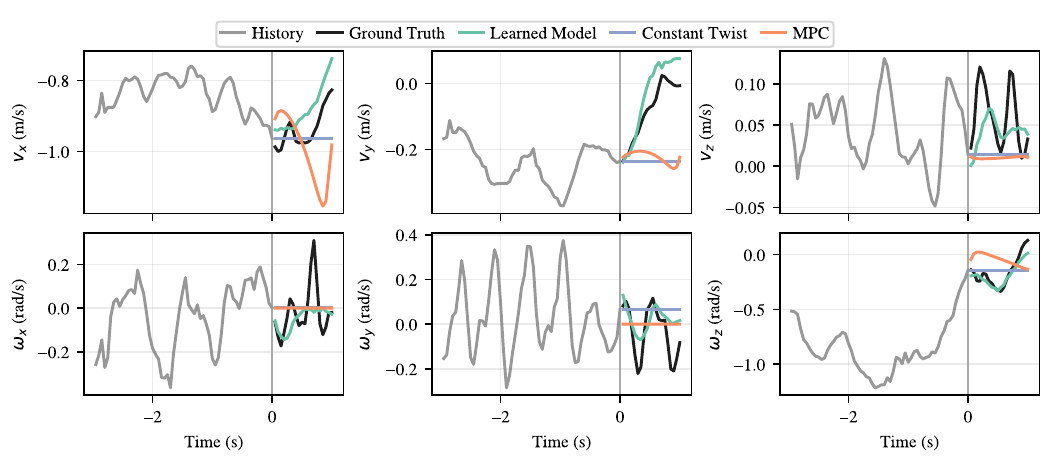}
        \caption{Predicted object twists.}
        \label{fig:model-predictions}
    \end{subfigure}

    \caption{
        Qualitative prediction performance on a test set window.
        The learned model more closely matches the future collaborative motion of the human dyad than MPC or constant twist.
        Top: object motion history and predicted future motion visualized from above.
        Bottom: corresponding predictions for all six twist components. The vertical line separates observed history from future ground truth and predictions.
    }

    \label{fig:model-qualitative}
\end{figure*}

\textbf{Results}.
We report mean absolute error (MAE) for various prediction lengths on the test set in \figref{fig:model-perf-test}.
Across prediction lengths, the learned model consistently achieves lower MAE than the constant-twist and MPC baselines.
Prediction error increases with horizon length for all algorithms, but the learned model's error grows more gradually, suggesting that it captures collaborative behavior more effectively.
The error in translational twist components is similar between the learned model and constant twist, suggesting that short-term translational motion is often well-approximated by the current velocity.
% particularly at prediction horizons $\le$\qty{1}{\second},
The learned model achieves greater error reductions over constant twist in rotational twist components.
The evaluation environment imposes paths that require direction changes and transitions between straight and curved motion.
Such transitions are not handled well by constant twist as it cannot anticipate changes in twist.
MPC exhibits comparatively large error in both components because its objective favors rapid goal progress and close tracking of the reference path without encoding interaction-dependent velocity variations.
\figref{fig:model-qualitative} illustrates these differences qualitatively during an example test window containing changes in speed and curvature.
By conditioning on recent object motion, the learned model better accounts for these variations.

\subsection{Pilot Study Experimental Setup}

\textbf{Controllers}.
% We compare three controllers, each integrating a different algorithm for generating object twist commands with the same admittance layer and whole-body controller.
% We compare three controllers, each combining a different algorithm for generating future object twists with the admittance layer and whole-body QP:
We compare three controllers that share the same admittance layer and whole-body QP but differ in how they generate the nominal end-effector twist:
\begin{itemize}
    \item \emph{Compliance-only}: sets $\twist^{\ee,\nominal}_{t+1}=\mathbf{0}$ at every step, yielding a common, compliant controller that only reacts to user-applied forces.
    \item \emph{MPC}: uses the MPC baseline from the prediction evaluation as a hand-designed motion planner.
    \item \emph{\name{}}: our framework, using predictions from the learned model to provide anticipatory assistance.
\end{itemize}

\begin{figure}[t]
    \centering
    \includegraphics[width=.97\linewidth]{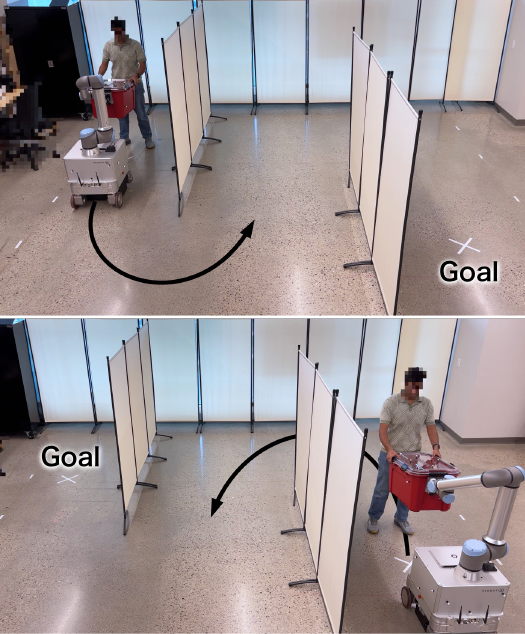}
    \caption{Starting configurations during the pilot study. Top: the robot starts ahead of the user along the direction of travel. Bottom: the robot starts behind the user.}
    \label{fig:pull-push}
\end{figure}

\textbf{Task Description}.
In each trial, the user and robot transport the object from marked start and goal locations in the constrained workspace while avoiding collisions with obstacles.
We evaluate the task in two configurations that reverse the robot's position relative to the direction of travel: one in which the robot is ahead of the user and another in which it is behind the user (\figref{fig:pull-push}).
Six users from our research group complete 18 trials each, with three trials per controller in each of the two configurations.
Before evaluation, all users, none of whom had prior experience with the system, are familiarized with the robot's compliance by applying forces to move the object independently along each translational and rotational degree of freedom.
Each user then completes two practice trials with an experimenter to learn the task procedure before beginning the robot trials.
% This is maybe not really important since we don't do detailed analysis on push vs pull configurations.
% The 18 robot trials are divided into three six-trial blocks.
% Within each block,
% Each controller is evaluated in two configurations: a pull configuration in which the robot is ahead of the user and a push configuration in which it is behind the user (\figref{fig:pull-push}).
% prev: "order within blocks is"
The controller order is randomized for each user, and users are not told which controller is active.
Similar to prior work, we keep object roll and pitch constant during trials to match typical dyadic behavior~\citep{lima2023assistive,shaw2025understanding}.

\textbf{Metrics}.
We compare controllers in terms of the following objective metrics:
\begin{itemize}
    \item \emph{Success Rate}: Proportion of interactions in which the object reaches the goal without collision or grasp loss.
    % \item \textbf{Number of Task Interruptions}: Number of contiguous time segments in which the linear speed of the object falls below a specific threshold. (\todo{define duration, speed thresholds}) (Pauses/interruptions rarely happened in the evaluation.)
    \item \emph{Time to Task Completion}: Duration from the object's first motion until it reaches the goal.
    \item \emph{Average Speed}: Mean magnitude of the object's translational velocity.
    \item \emph{Average Acceleration}: Mean magnitude of the object's translational acceleration.
    % \item \textbf{Average Jerk}: Mean magnitude of the object's translational jerk.
    \item \emph{Average Force and Torque}: Mean force and torque magnitudes measured at the robot end effector.
    % Work would measure zero even if the human is pushing/pulling hard while the object is stationary.
    \item \emph{Interaction Work}: $\int \lvert \wrench_t^\ee \cdot \twist_t^\ee \rvert\,dt$, used to quantify physical interaction effort during the task.
    % \item Interaction Force? I need to re-check how prior work comes up with $F_{reference}$ for the $F_{interaction} = F_{reference} - F_{measured}$ calculation.
\end{itemize}
% We compute speed, acceleration, and jerk after averaging object twists within non-overlapping \qty{50}{\milli\second} windows to reduce high-frequency noise when computing numerical derivatives.
We compute speed and acceleration after averaging object twists within non-overlapping \qty{50}{\milli\second} windows to reduce high-frequency noise when computing numerical derivatives.

\textbf{Implementation}.
We use a Neobotix MMO-700 mobile manipulator with an omnidirectional base, a 6-DoF Universal Robots UR10e arm, and an OnRobot VG10 suction cup gripper (\figref{fig:setting}).
We use the UR10e's built-in end-effector force-torque sensor.
Due to noise and limited precision of the UR10e force-torque sensor, we instantiate the low-pass filter for $\tilde{\wrench}^{\ee}_t$ (eq.~\eqref{eq:admittance-update}) as a first-order filter with a cutoff frequency of \qty{0.8}{\hertz} and deadbands of \qty{5}{\newton} on force components and \qty{1}{\newton\meter} on torque components, and zero the force-torque sensor before each trial.
We tune admittance parameters to balance smoothness and responsiveness during isolated movements of the object along each degree of freedom:
\begin{align*}
    M &= \operatorname{diag}(2.0, 2.0, 2.0, 0.05, 0.05, 0.05), \\
    D &= \operatorname{diag}(25.0, 25.0, 25.0, 6.0, 6.0, 6.0).
\end{align*}
We use the following parameters for the QP:
\begin{align*}
    W_u &= \operatorname{diag}(0.1, 0.1, 0.1, 0.1, 0.1, 0.1, 0.3, 0.3, 0.3) \\
    W_s &= \operatorname{diag}(8.0, 8.0, 8.0, 8.0, 8.0, 8.0, 30.0, 30.0, 10.0) \\
    W_p &= \operatorname{diag}(0.75, 0.75, 0.75, 0.75, 0.75, 0.67) \\
    W_\varepsilon &= \operatorname{diag}(30.0, 30.0, 30.0, 15.0, 15.0, 15.0) \\
    w_\alpha &= 2.0 \\
    u_{\max} &= (0.75, 1.0, 1.0, 1.0, 1.0, 1.0, 0.8, 0.8, 1.5) \\
    u_{\min} &= -u_{\max} \\
    \varepsilon_{\max} &= (0.015, 0.015, 0.015, 0.05, 0.05, 0.05) \\
    \kappa &= 0.5
\end{align*}
% These limits were a bit arbitrary. The base has a maximum translational speed of 0.94 m/s based on wheel motors, but at that speed the robot cannot turn. The translational limit 0.8 m/s was chosen to be high enough to enable "realistic" interaction but it wasn't rigorously tuned.
% We could use a safety argument, but there isn't really a consensus on what safe limits should be (ISO/other standards are to conduct application-specific risk-assesments per my understanding). At the acceneration limits of the robot, the velocity limits have correspond to a maximum stopping time of \qty{500}{\milli\second} and a max stopping distance of \qty{400}{\milli\meter} during normal operation.
We limit the translational and rotational speeds of the base to \qty{0.8}{\meter\per\second} and \qty{1.5}{\radian\per\second}, respectively.
An experimenter continuously monitors each trial while holding a wireless emergency stop device.
The deployed learned model uses a \qty{3}{\second} history and a \qty{1}{\second} prediction horizon to balance future anticipation and real-time execution.
We use the same deterministic reference path across evaluation trials to isolate differences in execution from variation in global planning.
MPC and \name{} run at \qty{20}{\hertz}, while the admittance layer and QP run at \qty{500}{\hertz} to match the UR10e control rate.
All controllers run in real time on an Intel i7-13700 CPU.

\begin{table*}[t]
    \centering
    \caption{Summary of real-world collaborative transport performance across 36 trials for each controller. Continuous metrics are reported as mean (SD) across participant-level means $(N=6)$ and exclude failed trials.}
    \label{tab:real-world-results}
    \resizebox{\textwidth}{!}{%
    \begin{tabular}{lrrrrrrr}
        \toprule
        Algorithm
        & Success Rate $\uparrow$
        & Time (\unit{\second}) $\downarrow$
        & Avg. Speed (\unit{\meter\per\second}) $\uparrow$
        & Avg. Accel. (\unit{\meter\per\second\squared}) $\downarrow$
        % & Avg. Jerk (\unit{\meter\per\second\cubed}) $\downarrow$
        & Avg. Force (\unit{\newton}) $\downarrow$
        & Avg. Torque (\unit{\newton\meter}) $\downarrow$
        & Int. Work (\unit{\joule}) $\downarrow$ \\
        \midrule
        Compliance-only
        & \textbf{36/36}
        & 20.1 (2.2)
        & 0.445 (0.062)
        & \textbf{0.50} (0.08)
        % & \textbf{9.44} (1.51)
        & 13.4 (1.5)
        & 3.06 (0.33)
        & 115.4 (17.9) \\
        MPC
        & 35/36
        & 18.8 (2.7)
        & 0.478 (0.071)
        & 0.84 (0.09)
        % & 13.30 (1.43)
        & 9.3 (1.7)
        & 2.56 (0.55)
        & 59.2 (13.3) \\
        \name{}
        & 35/36
        & \textbf{17.5} (2.2)
        & \textbf{0.520} (0.068)
        & 0.84 (0.27)
        % & 12.34 (3.42)
        & \textbf{8.2} (1.5)
        & \textbf{2.53} (0.77)
        & \textbf{47.1} (16.6)\\
        \bottomrule
    \end{tabular}%
    }
\end{table*}

% \begin{figure*}[t]
%     \centering
%     \includegraphics[width=\textwidth]{figures/boxplots.pdf}
%     \caption{Per-trial metric distributions across controllers in real-world trials. Boxes show interquartile ranges, center lines show medians, and whiskers show 5th--95th percentiles.}
%     \label{fig:metrics-boxplots}
% \end{figure*}

\begin{figure*}[t]
    \centering
    \includegraphics[width=\textwidth]{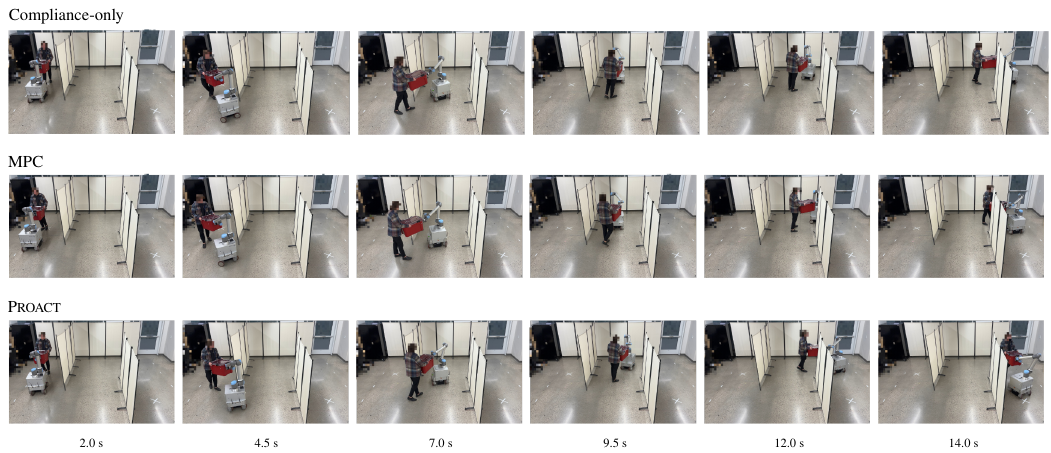}
    \caption{Snapshots from our pilot study, comparing Compliance-only, MPC, and \name{} across matching timestamps during trials.
    \name{}'s anticipation of collaborative behavior enables the team to progress more efficiently while requiring less interaction effort.
    }
    \label{fig:qualitative-real}
\end{figure*}

\subsection{Pilot Study Results}

\textbf{Aggregate performance}. \tabref{tab:real-world-results} summarizes real-world performance across 108 trials.
All three controllers achieve comparable success rates, with MPC and \name{} each encountering a single failure due to a suction grasp break before reaching the goal.
\name{} achieves the highest mean transport speed and, among successful trials, the shortest mean completion time: 12.9\% shorter than Compliance-only and 6.9\% shorter than MPC.
Notably, this faster transport is accompanied by the lowest mean interaction force, torque, and work.
Relative to Compliance-only, these quantities decrease by 38.8\%, 17.3\%, and 59.2\%, respectively.
Relative to MPC, they decrease by 11.8\%, 1.2\%, and 20.4\%.
Unlike Compliance-only, both MPC and \name{} contribute active, task-directed motion, reducing reliance on user-applied physical input to drive the interaction.
Together with the model's lower prediction error, \name{}'s lower interaction force and work suggest that its learned predictions better capture collaborative motion than MPC.

\textbf{Impact of starting configurations}. Examining the two starting configurations (\figref{fig:pull-push}) separately, we find that the controller ranking within each configuration is consistent with the aggregate results: \name{} achieves the lowest completion time, force, torque, and interaction work, while Compliance-only produces the lowest acceleration. % and jerk.
% On average across controllers, the push configuration, in which the robot is behind the user, reduced completion time, acceleration, jerk, force, and torque by 7.0\%, 17.7\%, 16.5\%, 6.4\%, and 20.0\%, respectively.
Across controllers, trials beginning with the robot behind the user have 7.0\%, 17.2\%, 5.3\%, 19.4\%, and 4.8\% lower completion time, acceleration, force, torque, and interaction work, respectively, than those beginning with the robot ahead.
A possible explanation is that positioning the robot at the trailing end reduces the user's uncertainty about the robot's intended motion.
Nevertheless, \name{}'s advantage over MPC and Compliance-only persists across both configurations.

\textbf{Implications of proactive behavior}. \name{} exhibits more proactive behavior compared to Compliance-only and MPC during changes in path curvature.
In particular, the robot had a tendency to get stuck in corners when running the Compliance-only controller, requiring the user to maneuver it out.
\figref{fig:qualitative-real} illustrates these qualitative differences, particularly between $t=\qty{7.0}{\second}$ and $t=\qty{12.0}{\second}$.
% However, \name{} also exhibited 66.0\% higher average acceleration and 30.6\% higher average jerk compared to the Compliance-only controller.
However, \name{} also exhibits 67.4\% higher average acceleration compared to the Compliance-only controller.
% We observe similar high average acceleration and jerk with the MPC controller.
We observe similarly high average acceleration with MPC.
Two factors may have contributed to this.
First, more proactive behavior through curves produces faster directional changes in the object's velocity, increasing the acceleration magnitude reported by the metric.
% Second, high-level predictions may propose changes in motion to be proactive, but they are not always in agreement with the user.
% \elvin{faster acceleration during orientation changes}
Second, the suction gripper's pliable silicone cups can introduce spurious wrench measurements.
Combined with the phase lag introduced by low-pass filtering, these measurements can produce an oscillatory admittance response.
Because the resulting object motion is subsequently observed by the predictive controllers, this oscillation may be reinforced through feedback.
We observed such oscillation before the grasp breaks in the failed MPC and \name{} trials.

% In addition, we collect the following subjective metrics via verbal interaction with the participant:
% \begin{itemize}
%     \item User Preferences: ``How would you rank your preferences between the different robot algorithms as a task partner? What if we include the human-human interaction in those preferences?''
%     \item User Impressions: ``Do you have additional thoughts about any particular algorithm?''
% \end{itemize}

% Presentation: what % of the time is the velocity prediction model judged as the best controller; what % of the time it "wins" against the other two controllers

\section{Discussion}

\textbf{Summary}. We presented \name{}, a framework combining collaborative behavior prediction with compliant whole-body control to deliver proactive assistance in physical HRC. Unlike prior work that emphasizes purely compliant systems, \name{} empowers a robot to act as a proactive partner rather than a reactive follower. We deployed \name{} on a 9-DoF mobile manipulator and demonstrated efficient human-robot collaborative transport on tasks of practical relevance, in terms of time, interaction forces, and interaction work in a constrained and obstacle-occupied workspace.

% The learned model represents collaborative behavior through the motion of the carried object.
% motivated by practical considerations during deployment and in available data,

\textbf{Limitations}. Our pilot study validated the superior functional performance of \name{} but did not capture subjective user perceptions. Ongoing work involves running a large-scale user study to extract user insights about interacting with our system. While our representation of collaborative behavior through object motion is motivated by available data and practical deployment considerations, it does not account for user ergonomics, verbal communication, or other cues that users may draw on when coordinating with a robot.
Future work will also investigate learning more general models of collaboration from additional signals and richer perception.
Our evaluation focused on a single, controlled setting to isolate the effects of different robot controllers during physical interaction.
We also assumed the user and robot have grasped the object and established a shared objective.
We aim to integrate our framework with higher-level decision-making and communication frameworks to reason about how users and robots should initiate and negotiate interactions within a shared vision-language context.
Finally, future work will target generalization across a wider range of object shapes, masses, and workspace layouts.

% The current implementation also relies on motion capture and filtered force-torque measurements
% More precise sensing or adaptive coupling between the learned model and the admittance layer, for example through shared-control or arbitration schemes, may further improve system performance.
% there is also the HRI angle of understanding how users felt about the interaction with the different algorithms that could be part of future work along the lines of \citet{yang2025implicit}

\balance
\bibliographystyle{IEEEtranN}
{
\footnotesize
\bibliography{references.bib}
}

% \newpage
% \appendix
% \input{sections/08-appendix}

\end{document}